\documentclass{article}
\usepackage{spconf,amsmath,graphicx}
\usepackage{arydshln}
\usepackage{multirow}
\usepackage{graphicx}
\usepackage{array}
\usepackage{multirow}
\usepackage{colortbl}
\usepackage{tabularx}
\usepackage{booktabs}
\usepackage{xcolor}
\usepackage{threeparttable}
\usepackage{booktabs}
\usepackage{amsmath}
\usepackage{amsfonts}
\usepackage{tabu}
\usepackage{color}
\usepackage{graphicx}
\usepackage{threeparttable}
\usepackage{colortbl}
\usepackage{bbding}
\usepackage{tabu}
\usepackage{graphicx}
\usepackage{booktabs}
\usepackage{multirow}
\usepackage{multirow}
\usepackage{multicol}
\usepackage{makecell}
\usepackage{booktabs}
\usepackage{algorithm}
\usepackage{algorithmic}
\usepackage{color}
\usepackage{amsmath,amssymb}
\usepackage{booktabs}
\usepackage{CJK} 
\definecolor{darkblue}{rgb}{0.0,0.0,0.55}
\definecolor{darkgray}{rgb}{0.66,0.66,0.66}

\title{Isochrony-Controlled Speech-to-Text Translation: A study on translating from Sino-Tibetan to Indo-European Languages}

\name{BLIND}
\address{BLIND}

\name{\begin{tabular}{c}
Midia Yousefi,
Yao Qian,
Junkun Chen,
Gang Wang,
Yanqing Liu,\\
Dongmei Wang,
Xiaofei Wang,
Jian Xue
\end{tabular}}
\address{Microsoft, One Microsoft Way, Redmond, WA, USA}

\begin{document}
%\ninept
%
\maketitle
\begin{abstract}
End-to-end speech translation (ST), which translates source language speech directly into target language text, has garnered significant attention in recent years. Many ST applications require strict length control to ensure that the translation duration matches the length of the source audio, including both speech and pause segments. Previous methods often controlled the number of words or characters generated by the Machine Translation model to approximate the source sentence's length without considering the isochrony of pauses and speech segments, as duration can vary between languages. To address this, we present improvements to the duration alignment component of our sequence-to-sequence ST model. Our method controls translation length by predicting the duration of speech and pauses in conjunction with the translation process. This is achieved by providing timing information to the decoder, ensuring it tracks the remaining duration for speech and pauses while generating the translation. The evaluation on the Zh-En test set of CoVoST 2, demonstrates that the proposed Isochrony-Controlled ST achieves $0.92$ speech overlap and $8.9$ BLEU, which has only a $1.4$ BLEU drop compared to the ST baseline.
\end{abstract}

\begin{figure*}[t]
\centering

\includegraphics[width=\linewidth]{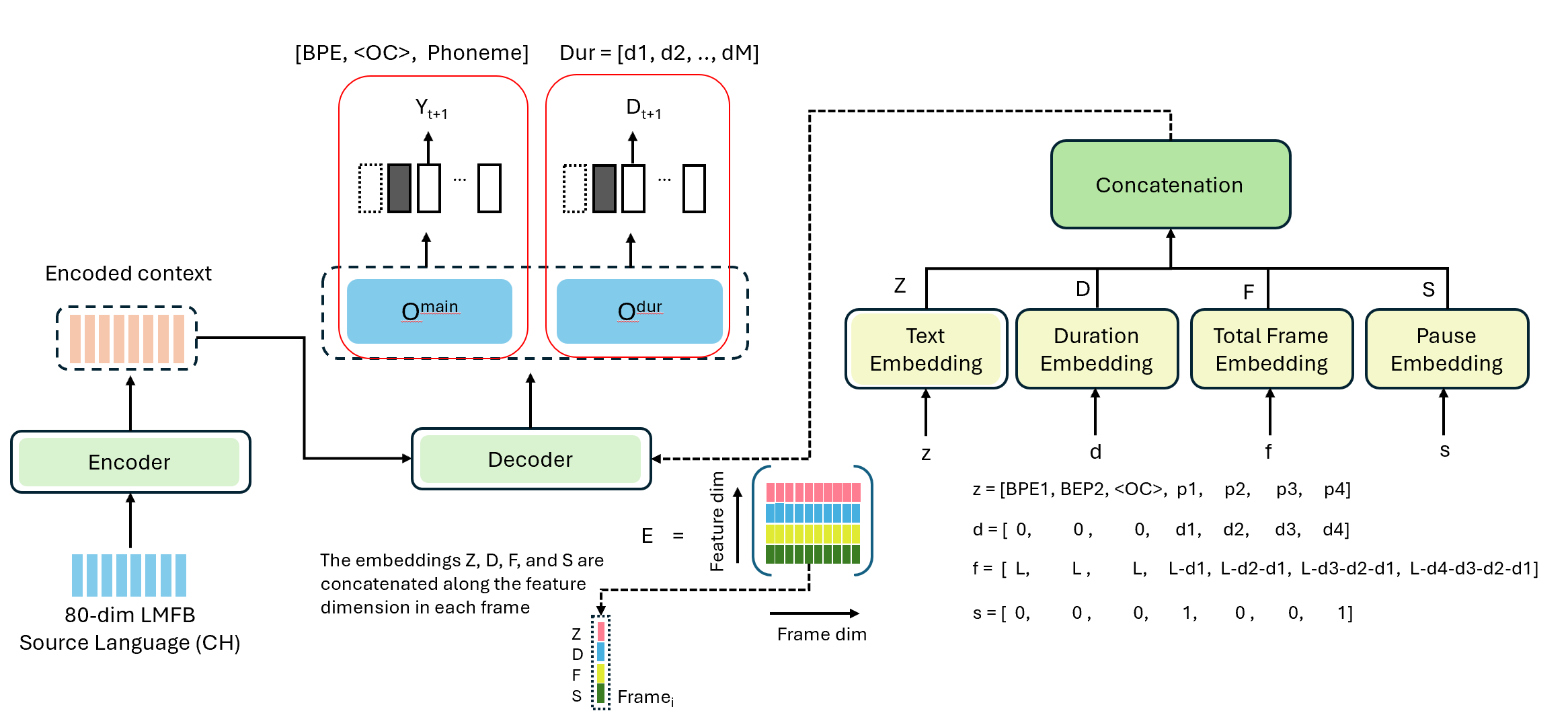}

\caption{The proposed Isochrony-controlled E2E Speech-to-Text Translation model.}
\label{fig:model}
\vspace{-0.2cm}
\end{figure*}

\section{Introduction}
\label{sec:intro}

Speech-to-text Translation (ST) is a technology that converts speech input in one language into text output in another language \cite{berard2016listen, chen2021specrec, vila2018end, ren2020simulspeech}. This technology has numerous real-world applications, such as simultaneous translation for international conferences \cite{ren2020simulspeech, ma2020simulmt,wang2023better}, enhancing accessibility \cite{bouillon2021speech, norre2022investigating} and communication across language barriers \cite{downing2002effective, bano2020speech}. Additionally, ST  can be integrated with Text-to-Speech Synthesis (TTS) \cite{taylor2009text, tan2021survey, kumar2023deep} to form a core component of cascaded Speech-to-Speech Translation systems \cite{lee2021textless, kundu2022survey}, which are particularly useful in video dubbing applications \cite{lakew2022isometric,federico2020speech}. In these applications, controlling the prosody alignment and duration of the generated translation is crucial \cite{virkar2022prosodic,wu2023videodubber}. In ST tasks, proper duration control is crucial to maintain synchronization between the source audio and the translated text or speech, ensuring that the translation is delivered within the same timeframe as the original speech. This synchronization is essential for applications like video dubbing, where the translated speech must match the timing of the speaker's lip movements \cite{patel2023visual}, and for captioning, where subtitles need to appear in sync with the audio \cite{virkar2021improvements, federico2020evaluating}. Therefore, precise control over prosody and duration enhances the overall quality and usability of ST in various multimedia applications \cite{tam2021isochrony,le2024transvip}. Conventional Speech-to-Text Translation (ST) methods typically involve a two-step process where Automatic Speech Recognition (ASR) \cite{kaur2021automatic, malik2021automatic} and Machine Translation (MT) \cite{dabre2020survey, stahlberg2020neural, rivera2022machine} are cascaded together. In these approaches, the responsibility for maintaining isochrony and the duration alignment between the source speech and the translated text falls on the MT module \cite{pal2023improving}. Their goal is to ensure that the translated text mirrors the timing of the original speech as closely as possible.

In \cite{lakew2022isometric,lakew2021machine}, the authors attempt to achieve isochrony in Machine Translation by using isometry, which means generating translations that match the number of characters in the source text. However, this approach has its own limitations. Research \cite{brannon2023dubbing} has shown that isometry is only weakly correlated with isochrony, meaning that matching character counts does not necessarily result in translations that align well with the timing of the source speech. An alternative approach, as discussed in \cite{chronopoulou2023jointly}, involves predicting the duration of each translated word in conjunction with generating the word sequences. By interleaving word duration predictions with the translated words, this method aims to achieve more precise control over the timing of the translation. This approach allows the translation to maintain better synchronization with the source speech, improving the overall quality and naturalness of the translated output. Expanding on this idea, authors of \cite{pal2023improving} explored target factors and auxiliary counters to predict duration sequences jointly with target language phoneme sequences. In their approach, target factors enabled the estimation of the phonemes and duration sequence separately while still ensuring that they are conditioned
on each other. However, the aforementioned approaches control the duration of the translated speech in the MT module in the cascaded ST approach, and their performance might be negatively affected by the error propagation of the ASR and Voice Activity Detection (VAD) modules.  

In this study, we aim to conduct Isochrony-Controlled Speech-to-Text Translation for Sino-Tibetan languages into English. The Sino-Tibetan and Indo-European language families are distinct, with the former predominantly featuring tonal, monosyllabic languages with simpler grammar, and the latter comprising non-tonal, polysyllabic languages with more complex grammar, which presents challenges in speech translation \cite{driemsino}. Additionally, the conciseness of Sino-Tibetan languages like Chinese, which are often two-thirds the length of their Indo-European counterparts such as English in written form, poses further difficulties in achieving isochrony in translation \cite{hoosain1983processing}. We propose a sequence-to-sequence ST system and specific pre-training data collection to ensure competitive performance on CoVoST2 \cite{wang2021covost} benchmark. Additionally, we propose to integrate embedded utterance timing information into the decoder to condition the translation onto the duration modeling for both speech and pause segments. Our evaluations on real and synthetic data shows in terms of both isochrony and translation quality over our sequence-to-sequence ST baseline.

The remainder of this paper is as follows. In section \ref{sec:method}, the proposed method is described. Section \ref{sec:experiments} explains the details of the data preparation and experiment setup. Finally, section \ref{sec:analysis} provides the analysis and conclusions of the proposed method.

\section{End-to-End Speech-to-Text Translation}
\label{sec:method}
\subsection{Training}
Given the source speech feature sequence $x=(x_1, x_2, ..., x_N)$ and the translation sequence $y = (y_1,y_2, ..., y_M)$ in target language, where each $x_i$
represents the frame-level feature with a certain duration and each $y_j$ is a text token or unit, the End-to-End ST (E2E ST) aims to maximize the likelihood of the translation sequence conditioned on acoustic features $x$ and model parameters $\theta$ as follows:
This objective can be expressed as follows:
\vspace{-3mm}
\begin{equation}
p_{\text{translation}}(y \mid x; \theta) = \prod_{t=1}^{N} p(y_t \mid x, y_{<t}; \theta),
\label{eq:opt}
\end{equation}
in which the probability of generating the correct translation $y$ given the source speech sequence $x$ and the model's parameters $\theta$ is optimized. In a conventional E2E ST, as shown in Eq.~\ref{eq:opt}, only the token sequence is estimated. In order to control the length of the predicted token sequence, the duration of the $y_j$ sequence needs to be provided to the decoder. Therefore, we obtain the phoneme alignment for each utterance which includes the phoneme list and their associated duration sequence. As depicted in Figure \ref{fig:model}, the phoneme sequence is concatenated to the token (BPE) sequence interleaved by a special token $\textless OC\textgreater$ indicating Output Change. The final tensor $z$ is then fed to the text embedding module. Additionally, the corresponding timing sequence $d$ is prepared by concatenating zero to the phoneme duration sequence. Please note that since we do not have access to the duration of the BPE sequence, we need to pad the phoneme duration with zero to match the length of tensor $z$. Two other important timing information are the total remaining frames $f$ and the pauses/silences $s$ at each time frame of the speech utterance which are obtained from the alignment results. This information will facilitate controlling the length of the translation through predicting the duration for each generated translation unit and the silence frames. The details of the translation encoder, decoder and duration control are outlined below.

\textbf{Encoder-} The primary role of the encoder is to transform the frame-level source acoustic speech features $x$  into a set of high-dimensional representations that encapsulate the essential information from the input. This involves capturing temporal dependencies and extracting meaningful features from the speech signal. In this study, we deploy a transformer-based encoder \cite{vaswani2017attention} which uses self-attention mechanisms to process the entire input sequence in parallel. This allows the model to capture long-range dependencies more effectively. The input sequence is first embedded into a higher-dimensional space and then passed through multiple layers of self-attention and feed-forward networks.

\begin{equation}
h =\text{TransformerEncoder}(x).
\label{eq:encoder}
\end{equation}
The encoder outputs a context vector $h$ that summarizes the input sequence. This context vector is then passed to the decoder, which generates the target sequence.

\textbf{Decoder-} The transformer-based decoder specifically is designed to generate the target sequence (e.g., translated text) by predicting each subsequent word in the sequence based on both the context vector and the previously generated tokens. The input to the decoder $E$ consists of: 

\begin{figure}
\begin{center}
\includegraphics[width=1.0\linewidth]{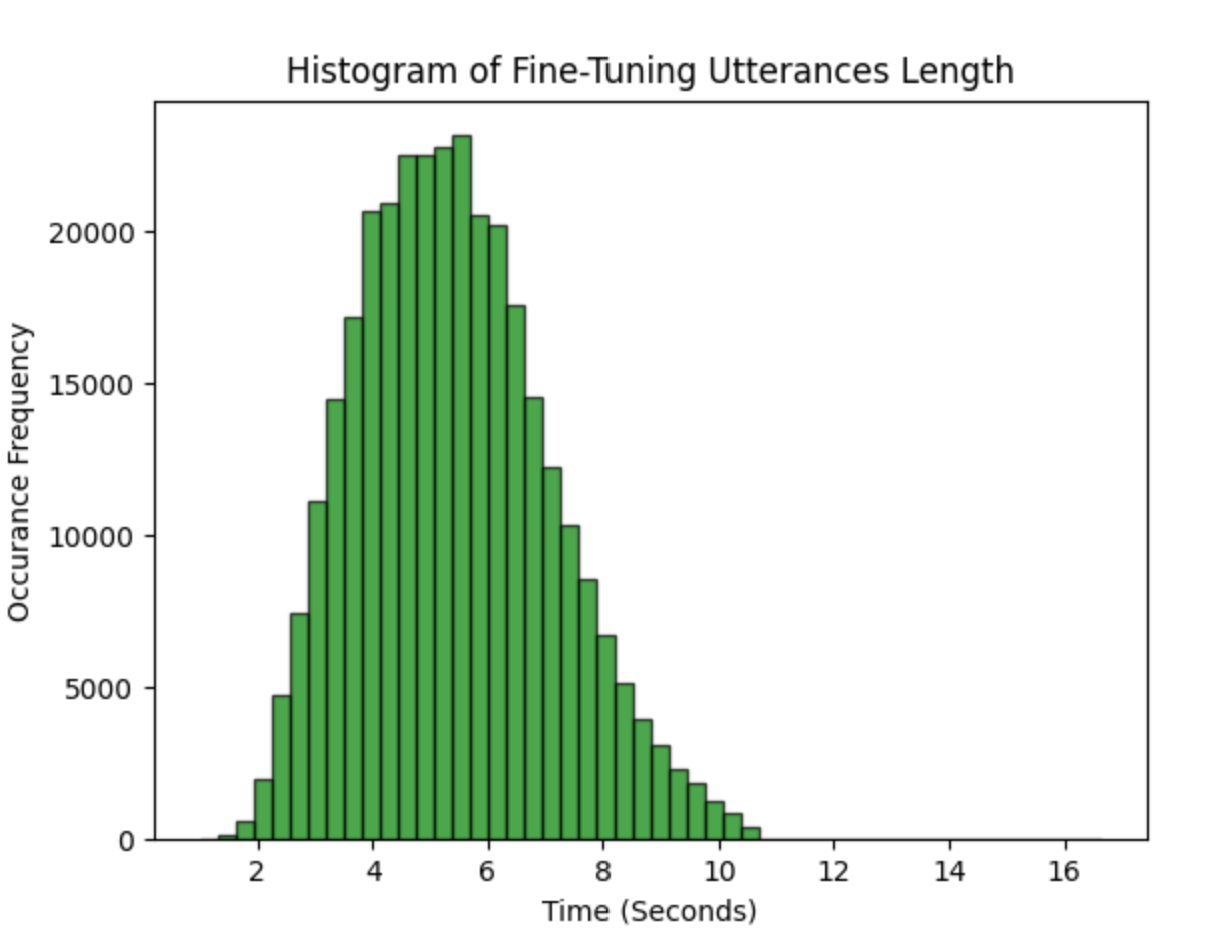}
 % \caption{Proposed sequence modeling block for multi-talker ASR with SD task.}

 \caption{Length histogram of the fine-tuning data.}
\label{fig:histogram}
\end{center}
\end{figure}

\vspace{-3mm}
\begin{align}
E &= \text{Concat}(Z, D, F, S),  \quad  E\in \mathbb{R}^{[B, T, 3\times I +I_z]} \label{eq:decoder_input} \\
Z &= \text{Embedding}(z) + \text{PosEncod}(z), \quad  Z\in \mathbb{R}^{[B, T, I_z]} \label{eq:decoder_input_z} \\ 
D &= \text{DurationEmbedding}(d), \quad D\in \mathbb{R}^{[B, T, I]} \label{eq:decoder_input_d} \\
F &= \text{TotalFramesEmbedding}(f), \quad F\in \mathbb{R}^{[B, T, I]} \label{eq:decoder_input_f} \\
S &= \text{PauseEmbedding}(s), \quad S \in \mathbb{R}^{[B, T, I]} \label{eq:decoder_input_s}
\end{align}
in which $Z$ is the text embedding, $D$ is the duration embedding, $F$ is the total remaining duration embedding, $S$ is the Silence embedding. In these equations, $B$ is the batch size, $T$ is the total number of frames, $I_z$ is the embedding dimension of text embedding $Z$, and $I$ is the embedding dimension of $D$, $F$ and $S$. Once the embeddings for translation sequence, duration, and total remaining frame and silences are calculated, they are concatenated along the feature dimension at each frame and are fed into the decoder to generate output $O$. Next, we have two linear layers: LinearToken and LinearDur to generate token and duration sequences respectively:

\begin{align}
O &=\text{TransformerDecoder}(E, h) \\
\hat{y}_{Tran} &= \text{LinearToken}(O_1), \quad  O_1=O[:,:,:I_z] \\
\hat{y}_{Dur} &= \text{LinearDur}(O_2), \quad  O_2=O[:,:,I_z:I_z + I]
\label{eq:decoder}
\end{align}
where $O_1$ is the first $I_z$ elements of the feature dimension across all frames in the tensor $O$. $O_2$ is the predicted duration features from$ I_z$ to $I_z + I$ in the tensor $O$. The model is updated by combining two loss functions:   
\emph{(i)} maximizing the KL-Divergence Loss between probability distributions of the generated translation $\hat{y}_{Tran}$ and the reference translation $y$;
% \jc{in my memory, we normally use entropy (label smoothing loss) for this.}
\emph{(ii)} the MSELoss between the estimated duration sequence $\hat{y}_{Dur}$ and the reference duration sequence $d$. 

 \begin{equation}
\text{Loss} =\text{KLDivLoss}(\hat{y}_{Tran}, y)  + \text{MSELoss}(\hat{y}_{Dur}, d).
\label{eq:encoder}
\end{equation}
\vspace{-5mm}

\renewcommand{\arraystretch}{1.5} % Adjust row height
\vspace{-1mm}
\begin{table}[t]
\caption{Statistics of pre-training speech corpora for each source language.}
\vspace{3mm}
\centering
\begin{tabular}{c c c c c}
\hline
Language &  & DE, ES, IT, FR &     &  EN, ZH \\
\hline
%Hour range per language &  & \textless 50k &   &  \textgreater 100k\\
Hours &  & \textless 50k &   &  \textless 100k\\
\hline
\end{tabular}
\label{tab:training_corpora}

\end{table}

\subsection{Inference}
During inference, the input to the decoder $E$ is initialized at decoding step $0$, and then calculated for the next decoding steps. Since $E$ is the concatenation of multiple embedding tensors, we initialize each tensor differently: $z$ is initialized by a special toke $\textless sos\textgreater$ which indicates the start of a sentence, $d$ is initialized by $0$ because the duration of the $\textless sos\textgreater$ is assumed to be $0$. The total number of frames of the source audio is used to initialize $f$, and finally $s$ is also initialized with $0$. These tensors are then fed into their corresponding embedding extraction module defined in Eq \ref{eq:decoder_input}-\ref{eq:decoder_input_s}. At each decoding step, based on the estimated phoneme duration $d_{t-1}$, we update the total remaining frames $f_t$, additionally, if the estimated token $z_{t-1}$ is predicted to be silence, we set $s_t$ to $1$ the next decoding step.  
\begin{table*}[t]
\centering
\begin{threeparttable}
\caption{The results of BLEU score on Zh-En subset of CoVoST2 test set. Our Baseline has a competitive performance compared to other studies conducted on this test set.VP-400k stands for VoxPopuli corpus, MLS is Multilingual LibriSpeech, CV stands for Common Voice, VL and BBL represent VoxLingua and BABEL datasets respectively. For the details of model architecture and training data, please refer to the original reference cited in column 1.
}
\label{tab:st_base}
\setlength\tabcolsep{4.8pt}
\begin{tabular}{c|c|c|c}

\toprule
Method                       & Architecture        &Pre-trained models  and Training data  &Score \\ \midrule
CoVoST 2 \cite{wang2021covost} & Transformer Encoder/Decoder         &Pre-trained English ASR encoder, CoVoSt2    & 5.8       \\

CVSS \cite{jia2022cvss} &  Translatotron 2 \cite{jia2022translatotron}          &Pre-trained multilingual ASR, CoVoST2 & 10.7       \\

XLS-R \cite{babu2021xls} & wav2vec 2.0 \cite{baevski2020wav2vec} + Transformer Decoder     2   & VP-400K, MLS, CV, VL, BBL, CoVoST2     & 9.4       \\

XMEF-X \cite{li2020multilingual} & wav2vec 2.0 \cite{baevski2020wav2vec} + mBART Decoder    &Pre-trained wav2vec, Europarl ST, CoVoST2     & 8.9       \\

\midrule

Our ST Baseline & Transformer Encoder/Decoder       & In-house multilingual ASR data, CoVoST2   & 10.3     \\\bottomrule
\end{tabular}
\end{threeparttable}
\end{table*}
we use Beam Search to obtain fluent translation. Beam search keeps track of the top $5$ most promising translations at each step, ensuring that the final translation is both fluent and accurate without having to consider every single possible translation.

\section{Experiments}
\label{sec:experiments}
\textbf{Data} -- we pre-train our proposed ST model using data prepared similar to   \cite{xue2023weakly, chen2023improving}. A large-scale in-house multi-lingual ASR dataset with hundreds of thousands hours of speech audio is used in the pre-training stage. The details of the pre-training data is shown in Table \ref{tab:training_corpora}. This data includes 6 high-resource languages with an ASR transcription.
%a total of 481k hours of speech data.
All the training data has been anonymized, with any personally identifiable information removed. A text-based machine translation service, such as GPT-4, is used to convert the ASR transcriptions into target language texts for ST training. The model is built with a direct translation framework, enabling it to process speech from a set of languages, specifically German (DE), English (EN), Spanish (ES), Italian (IT), French (FR), and Chinese (ZH), into English (EN) seamlessly, without requiring any language-specific configurations.

During the fine-tuning stage, we utilized the Common Voice corpus \cite{ardila2019common}. Specifically, we focused on English audio files and their corresponding Chinese transcriptions. Leveraging a zero-shot text-to-speech synthesizer, we generated Chinese utterances using the Chinese text and English audio prompts. This process enabled us to create a comprehensive dataset consisting of 400 hours of paired English-Chinese audio and text data. The majority of the utterances in this data are less than 10 seconds long. The length histogram of the data is plotted in Figure \ref{fig:histogram}. We extract 80-dimensional log mel-scale filter
bank features (windows with 25ms size and 10ms shift). We use a subword tokenizer with a vocabulary size of 5k, trained using SentencePiece \cite{kudo2018sentencepiece}.

For our evaluation, we use the Zh-En test set from the CoVoST2 corpus \cite{wang2021covost}. Additionally, we set aside 2,000 utterances from the synthesized speech generated for fine-tuning the ST model as a secondary test set. The purpose of using a synthesized test set is to address the distribution mismatch between the synthesized fine-tuning data and the real recordings of the Zh-En CoVoST2 test set, which might negatively impact translation quality and length prediction. Therefore, besides assessing performance on CoVoST, we also evaluate our model on synthesized speech to demonstrate the proposed method's capability in handling train/test distribution mismatches.

\textbf{Model} -- we use sequence-to-sequence transformer architecture \cite{vaswani2017attention} for both encoder and the decoder modules. The 80-dim Log Mel-Filterbanks (LMFB) are fed into $4$ layers of 2-d Convolutional layers with Kernel size of $(3, 3)$ and stride of $(1, 1)$  with channels size of $64, 64, 128, 128$ respectively. We apply also layer normalization between each convolutional layer. The encoder consists of $18$ layers of Conformers, with $8$ attention head and $512$ attention dimension, and $10\%$ dropout rate.

The decoder consists of embedding layers and a $6$-layer Transformer decoder module with $8$ attention head each of which has $512$ feature dimension. The DurationEmbedding, TotalDurationEmbedding, and PauseEmbedding all have the same architecture: consisting of a 512-dim linear layer followed by layer normalization and RelU activation function. 
The last block of the decoder includes two linear layers to produce two output sequences: \emph{(i)} a linear layer with a dimension size of $4366$ which is the size of vocabulary list and generates the logits for the translation hypothesis, and \emph{(ii)} a linear layer with output dimension $1$ to produce the duration sequence. All our models are trained using AdamW optimizer with a linear decay learning rate of $5e-5$ and batch size of 10k frames for 5M steps on 8 GPU machines. 

\textbf{Evaluation metrics} -- we evaluate the performance of our models from two aspects, translation quality and translation length. The translation quality is measured using the sacreBLEU\footnote{https://github.com/mjpost/sacreBLEU} \cite{post2018call}. Translation length is measured by computing speech overlap obtained from the estimated total number of frames. The speech overlap is calculated as:

\begin{equation}
\text{Speech Overlap} = 1 - \frac{|\text{ref. duration} - \text{hyp. duration}|}{\text{ref. duration}}.
\label{eq:speech_ovr}
\end{equation}
constraining the length of the generated translation imposes a trade-off between the length and the quality of the translation. In applications such as video-dubbing, it is important to compromise on the translation accuracy to ensure the translated speech still fits within the time frame of the original speech.

\begin{table}[t]
\centering

\caption{The results of BLEU score and speech overlap on Zh-En CoVoST2 test and synthesized Zh-En test sets. I-C ST represents Isochrony-Controlled Speech Translation in which the Duration Embedding, Total Frame Embedding and Pause Embedding have different fetaure dimension.}
\vspace{2mm}
\label{tab:results-LCST}
\begin{tabular}{lcccc}
\toprule
\multirow{2}{*}{Model}  & \multicolumn{2}{c}{Zh-En CoVoST2} & \multicolumn{2}{c}{Synthesized Zh-En} \\ 
\cmidrule(lr){2-3} \cmidrule(lr){4-5}
 & BLEU &Ovr & BLEU & Ovr \\ 
 \midrule
ST Baseline & 10.3 & - & 30.2 & - \\ 
\midrule
16-dim I-C ST & 7.8 & 0.91 & 27.8 & 0.80 \\ 
32-dim I-C ST & 7.0 & 0.62 & 26.1 & 0.50 \\ 
64-dim I-C ST & 8.6 & 0.48 & 27.8 & 0.38 \\ 
128-dim I-C ST & 8.4 & 0.56 & 26.8 & 0.53 \\ 
256-dim I-C ST & 8.9 & 0.83 & 27.1 & 0.57 \\ 
512-dim I-C ST& \textbf{8.9} & \textbf{0.92} & \textbf{28.7} & \textbf{0.89} \\ 
\bottomrule
\vspace{-10mm}
\end{tabular}
\end{table}

\section{Results and analysis}
\label{sec:analysis}

The results of our E2E ST baseline on Zh-En subset of CoVoST2 are shown in the last row of Table \ref{tab:st_base}. We have also reported the results of recent studies on the same test set in this table. However, please note that these results are not directly comparable due to different data used in pre-training, and different architectures. The motivation for providing this table is to show that our baseline has a competitive performance compared to the other recent approaches in the literature. Additionally, as mentioned in \cite{jia2022cvss} it is worth noting that in the CoVoST 2 corpus, the BLEU score is significantly lower on certain languages (e.g. zh) than others for two reasons: \emph{(i)} because in theses language the data include a lot of non-English names and proper nouns, which cannot be recognized/translated correctly, \emph{(ii)} Zh-En is considered a low-resource language in CoVoST2 corpus with only 10h of training data, compared to languages such as French and German with 180h and 119h of training data respectively. 
\begin{table}[!b]
\centering
\vspace{-7mm}
\caption{Variation of translation quality (BLEU) and speech
overlap with different amounts of noise added to the phoneme
duration sequence. We use the 512-dim I-C ST system to evaluate the effect of noisy duration on both test sets.}
\vspace{2mm}
\label{tab:results-noise}
\begin{tabular}{lcccc}
\toprule
\multirow{2}{*}{Model}  & \multicolumn{2}{c}{Zh-En CoVoST2} & \multicolumn{2}{c}{Synthesized Zh-En} \\ 
\cmidrule(lr){2-3} \cmidrule(lr){4-5}
 & BLEU &Ovr & BLEU & Ovr \\ 
 \midrule
512-dim I-C ST& 8.9 & 0.92 & 28.7 & 0.89 \\
\midrule
 + $\mathcal{N}(0, 0.01)$ &8.5 &0.75	 &28.1 & 0.84 \\
+$\mathcal{N}(0, 0.05)$ &8.9 &0.40	 &27.9 & 0.38 \\
+$\mathcal{N}(0, 0.1)$ &8.6 &0.62	 &28.7 & 0.59 \\
+$\mathcal{N}(0, 0.5)$ &8.9 &0.72	 &28.6 & 0.54 \\
+$\mathcal{N}(0, 1.0)$ &8.4 &0.95	 &28.5 & 0.92 \\
+ $\mathcal{N}(0, 1.5)$ &9.0 &0.72	 &29.22 & 0.80 \\

\bottomrule
\end{tabular}
\end{table}

\begin{CJK*}{UTF8}{gbsn}

\begin{figure*}[t]\centering
\small

    \resizebox{\linewidth}{!}{
    %\centering
    \small
    \setlength{\tabcolsep}{1.5pt}
    \renewcommand{\arraystretch}{1.}
    \begin{tabu}{p{3cm}|p{4cm}|p{4cm}|p{4cm}}
    \rowfont{\small}
    \toprule
     Source Transcription & Ground-Truth Reference & Isochrony-Controlled ST & Baseline ST\\
    \midrule
    一男一女在黄色花坛前摆姿势照相 &A man and a woman are posing for a photograph in front of a yellow flowerbed.  &A man and a woman pose for a picture in front of a yellow floral &A man and a woman posing a picture in front of a yellow flower position. \\
    % \midrule
    % 她的访客们通过这个论坛得到了许多问题的答案. & The forum allowed her visitors to get answers to the many questions they had. & His visitors received many questions through this forum. & His visitors received a number of questions through this forum. \\
    
    \midrule
    你说你从未恋爱过？ & You said that you had never been in love? & You said you never loved. & You said you were never in love \\
    
    \midrule

    当他还是个孩子的时候，他就学会了如何弹按钮式手风琴 & When he was a child, he learned how to play the button accordion & As a child, she learned how to play their accordion & When he was a child, she learned how to play buttons on the accordion \\
    
    \bottomrule
    \end{tabu}
    }
    
\caption{Decoding examples of the proposed Isochrony-Controlled ST and Baseline Translation. The first column shows the provided Chinese utterance transcription. The second column is the ground truth translation, and the third and fourth columns are the generated translations by the Isochrony-Controlled ST and Baseline Translation, respectively. Please note that the BLEU score for both the Isochrony-Controlled and baseline ST models range from 8.9 to 10.3, therefore, there are errors in the Translations in both scenarios.}
\vspace{-4mm}     
\label{fig:data_sample}     

\end{figure*}
\end{CJK*}

Table \ref{tab:results-LCST} presents the BLEU scores and speech overlap (Ovr) results for different models on the Zh-En CoVoST2 test sets and synthesized Zh-En test sets. The models include a baseline Speech Translation (ST) and various configurations of Isochrony-Controlled Speech Translation (I-C ST) with different duration dimensions $I:$ $16$, $32$, $64$, $128$, $256$, and $512$. The baseline model achieves the highest BLEU scores for both test sets, indicating the best translation quality without Isochrony control. However, as noted in \cite{pal2023improving}, there is a trade-off between translation quality and translation duration. Adding the Isochrony-control component preserves speech duration overlap but degrades translation quality. Among the Isochrony-Controlled Speech Translation (I-C ST) models, the 512-dim I-C ST achieves the best speech overlap percentage with the smallest BLEU score drop. On the Zh-En CoVoST2 test set, the 512-dim I-C ST achieves a 0.92 speech overlap with an 8.9 BLEU score, indicating a 1.4 BLEU drop. Despite this, compared to other systems in Table \ref{tab:st_base}, the proposed I-C ST still performs competitively in terms of translation quality compared to other recent studies. The same pattern is observed on the synthesized Zh-En test set, where the 512-dim duration embedding in the I-C ST model leads to a 0.89 speech overlap and a 28.7 BLEU score with only a 1.5 BLEU drop compared to the ST Baseline which has a BLEU score of $30.2$. One observation from Table \ref{tab:results-LCST} is the high variance in speech overlap (Ovr) results, which range from 0.48 to 0.92 on the Zh-En CoVoST2 test set and from 0.38 to 0.89 on the synthesized Zh-En test set. This variation is due to the fact that, as shown in Figure \ref{fig:histogram}, most of the utterances in our training and evaluation data are only a few seconds long. Consequently, generating even a single extra word in the translation can increase the audio length, leading to a drop in the speech Ovr metric.

We experiment with adding noise to the frame duration from the alignment in the training data to make the model more flexible with timing information. Results are shown in Table \ref{tab:results-noise}. The models are trained by adding Gaussian noise with different variances to the 512-dim I-C ST model, which performed best in Table \ref{tab:results-LCST}. Adding Gaussian noise with variance $1$ yields the best speech overlap (Ovr) for both test sets: $0.95$ for Zh-En CoVoST2 and $0.92$ for the synthesized Zh-En set, showing a $0.3$ improvement on both sets. However, this increased speech Ovr resulted in a BLEU drop of $0.5$ on CoVoST2 and $0.2$ on synthesized Zh-En, which is expected due to the short-length utterances in the training and testing sets.

Figure \ref{fig:data_sample} presents examples of translations generated by our proposed I-C ST model. These short examples pose a challenge for the ST model to produce high-quality yet concise translations. In the first example, the Isochrony-Controlled ST generates shorter and more accurate translation. In the second example, clearly a shorter translation is generated to satisfy the length requirement at the expense of lower quality translation. The same observations can also be made for the last examples where shorter translation are generated with little compromise on the quality and overall meaning of the sentence. Finally, based on the objective metrics and examples presented, our proposed model effectively balances the trade-off between translation quality and the speech overlap.

\section{Conclusions}
\label{sec:conclusions}

In this study we proposed Isochrony-Controlled Speech Translation (I-C ST)  model for Chinese-English language pair. Our proposed improvements to the duration alignment component of the sequence-to-sequence speech translation (ST) model have proven effective. By incorporating timing information directly into the decoder, we have enabled more precise control over translation length, ensuring that the duration of the translated speech and pauses aligns with the source audio. Our evaluation on the Zh-En subset of CoVoST2 demonstrates that these enhancements improve isochrony without compromising translation quality. This approach addresses the limitations of conventional ST baseline, which does not adequately consider the isochrony of pauses and speech segments, thus providing a more robust solution for ST applications requiring strict length control. The proposed method achieves a $0.92$ speech overlap with an $8.9$ BLEU score on the Zh-En CoVoST2 test set, which has only a $1.4$ BLEU drop compared to a conventional ST model. 


\bibliographystyle{IEEEbib}
\bibliography{main}

\end{document}